# Off-Line Handwritten Signature Retrieval using Curvelet Transforms

M. S. Shirdhonkar
Dept. of Computer Science and Engineering,
B.L.D.E.A's College of Engineering and Technology
Bijapur, India
E-mail:ms_shirdhonkar@rediffmail.com

Manesh Kokare
Dept. of Electronics and Telecommunication,
S.G.G.S Institute of Engineering and Technology
Nanded, India
E-mail: mbkokare@sggs.ac.in

*Abstract*—— In this paper, a new method for offline handwritten signature retrieval is based on curvelet transform is proposed. Many applications in image processing require similarity retrieval of an image from a large collection of images. In such cases, image indexing becomes important for efficient organization and retrieval of images. This paper addresses this issue in the context of a database of handwritten signature images and describes a system for similarity retrieval. The proposed system uses a curvelet based texture features extraction .The performance of the system has been tested with an image database of 180 signatures. The results obtained indicate that the proposed system is able to identify signatures with great with accuracy even when a part of a signature is missing.

I. INTRODUCTION (HEADING 1)

A. Motivation

A signature appears on many types of documents such as bank cheques in daily life and credit slips, thus signature has a great importance in a person's life. Automatic bank cheque processing is an active topic in the field of document analysis and processing. Signature validity confirmation of different document is one of the important problems in automatic document processing. Now a days, person identification and verification are very important in security and resource access control. For this purpose the first and simple way is to use Personal Identification Number (PIN), but PIN code may be forgotten. Now an interesting method to identification and verification is biometric approach [1]. Biometric is a measure for identification that is unique for each person. Always biometric is together with person and cannot be forgotten. In addition biometric usually cannot be misused.

Handwritten signature retrieval is still a challenging work in the situations of a large database. Unlike fingerprint palm print and iris, signatures have significant amount of intra class variations making the research even more compelling. This approach with the potential applications of signature recognition/verification system optimized with efficient signature retrieval mechanism.

B. Related works.

Signature verification contain two areas: off-line signature verification ,where signature samples are scanned into image representation and on-line signature verification, where signature samples are collected from a digitizing tablet which is capable of pen movements during the writing .In our work, we survey the offline signature identification and retrieval . In 2009, Ghandali and Moghaddam have proposed an off-line Persians signature identification and verification based on Image registration, DWT (Discrete Wavelet Transform) and fusion. They used DWT for features extraction and Euclidean distance for comparing features. It is language dependent method [1]. In 2008, Larkins and Mayo have introduced a person dependent off-line signature verification method that is based on Adaptive Feature Threshold (AFT) [2]. AFT enhances the method of converting a simple feature of signature to binary feature vector to improve its representative similarity with training signatures. They have used combination of spatial pyramid and equimass sampling grids to improve representation of a signature based on gradient direction. In classification phase, they used DWT and graph matching methods. In another work, Ramachandra et al [3], have proposed cross-validation for graph matching based off-line signature verification (CSMOSV) algorithm in which graph matching compares signatures and the Euclidean distance measures the dissimilarity between signatures.
In 2007, Kovari et. al presented an approach for off-line signature verification, which was able to preserve and take usage of semantic information[4].They used position and direction of endpoints in features extraction phase. Porwik [5] introduced a three stages method for offline signature recognition. In this approach the hough transform ,center of gravity and horizontal-vertical signature histogram have been employed, using both static and dynamic features that were processed by DWT has been addressed in[6].The verification phase of this method is based on fuzzy net using the enhanced version of the MDF(Modified Direction feature)extractor has been presented by Armand et.al [7].The different neural classifier such as Resilient Back Propagation(RBP), Neural network and Radial Basis Function(RBF) network have been used in verification phase of this method. In 1995, Han and Sethi [8], described offline signature retrieval and use a set of geometrical and topological features to map a signature onto 2D strings. We have proposed an offline signature retrieval model based on global features.
The main contribution of this paper is that, we have proposed off-line handwritten signature retrieval using curvelet transform, In retrieval phases Canberra distance measure is used. The experimental results of proposed method were satisfactory and found that it had better results compare with related works. The rest of paper is organized as follows: In section II, discusses the feature extraction phase. The signature retrieval is presented in section III. In section IV, the experimental results and finally section V concludes the work.

## II. FEATURE EXTRACTION PHASE

The major task of feature extraction is to reduce image data to much smaller amount of data which represents the important characteristic of the image. In signature retrieval, edge information is very important in characterizing signature properties. Therefore we proposed to use the curvelet transform. The performance of the system is compared with standard discrete wavelet transform which captures information in only three directions.

### A. Discrete Wavelet Transform

The multi resolution wavelet transform decomposes a signal into low pass and high pass information. The low pass information represents a smoothed version and the main body of the original data. The high pass information represents data of sharper variations and details. Discrete Wavelet Transform decomposes the image into four sub-images when one level of decomposing is used. One of these sub-images is a smoothed version of the original image corresponding to the low pass information and the other three ones are high pass information that represents the horizontal, vertical and diagonal edges of the image respectively. When two images are similar, their difference would be existed in high-frequency information. A DWT with N decomposition levels has 3N+1 frequency bands with 3N high-frequency bands [9], [10]. The impulse responses associated with 2-D discrete wavelet transform are illustrated in Fig. 1 as gray-scale image.

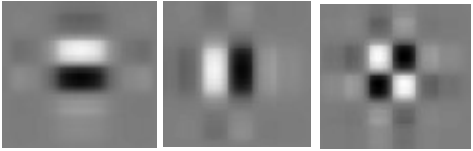

Fig. 1.Impulse response of $0^0$, $90^0$ and $\pm 45^0$ of DWT

### B. Curvelet Transform

Recently, Candµes and Donoho developed a new multiscale transform which they called the curvelet transform. Motivated by the needs of image analysis, it was nevertheless first proposed in the context of objects f(x1, x2) defined on the continuum plane (x1, x2) € R $^2$.

The transform was designed to represent edges and other singularities along curves much more efficiently than traditional transforms, i.e. using many fewer coefficients for a given accuracy of reconstruction. Roughly speaking, to represent an edge to squared error 1/N requires 1/N wavelets. The curvelet transform, like the wavelet transform, is a multiscale transform, with frame elements indexed by scale and location parameters. Unlike the wavelet transform, it has directional parameters, and the curvelet pyramid contains elements with a very high degree of directional specificity. In addition, the curvelet transform is based on a certain anisotropic scaling principle which is quite different from the isotropic scaling of wavelets. The elements obey a special scaling law, where the length of the support of a frame elements and the width of the support are linked by the relation width ≈ length$^2$.see details in [11].

### C. Feature Database Creation

To construct the feature vectors of each handwritten signature in the database using DWT and curvelet transform respectively. The Energy and Standard Deviation (STD) were computed separately on each sub band and the feature vector was formed using these two parameter values. The Energy $E_k$ and Standard Deviation $\sigma_k$ of k$^{th}$ sub band is computed as follows

$$E_k = \frac{1}{M \times N} \sum_{i=1}^{M} \sum_{j=1}^{N} |W_k(i,j)| \quad (1)$$

$$\sigma_k = \left[ \frac{1}{M \times N} \sum_{i=1}^{N} \sum_{j=1}^{M} (W_k(i,j) - \mu_k)^2 \right]^{\frac{1}{2}} \quad (2)$$

Where $W_k(i,j)$ is the $k^{th}$ wavelet-decomposed sub band, $M \times N$ is the size of wavelet decomposed sub band, and $\mu_k$ is the mean of the $k^{th}$ sub band. The resulting feature vector using energy and standard deviation are $\bar{f}_E = [E_1 \; E_2 \; ... \; E_n]$ and $\bar{f}_\sigma = [\sigma_1 \; \sigma_2 \; ... \; \sigma_n]$ respectively. So combined feature vector is
$$\bar{f}_{\sigma\mu} = [\sigma_1 \; \sigma_2 \; ... \; \sigma_n \; E_1 \; E_2 \; ... \; E_n] \quad (3)$$

## III. OFFLINE Handwritten SIGNATURE Retrieval PHASE

There are several ways to work out the distance between two points in multidimensional space. The most commonly used is the Canberra distance measure. It can be considered the shortest distance between two points. We have used Canberra distance metric as similarity measure. If x and y are the feature vectors of the database and query signature, respectively, x and y have dimension d, then the Canberra distance is given by

$$\text{Canb}(x, y) = \sum_{i=1}^{d} \frac{|x_i - y_i|}{|x_i| + |y_i|} \quad (4)$$

Algorithm 1: Offline Handwritten Signature Retrieval

Input: Test signature: St

    Feature database: FV

Output: Distance vector: Dist

    Handwritten signature retrieval

Begin

    Calculate feature vector of test signature using DWT and curvelet transform

  For each fv in FV do

    Dist= Calculate distance between test signature and fv using (4)

    sort Dist

  End for

    Display the top signature from dist vector.

End

## IV. EXPERIMENTAL RESULTS

### A. Image Database

The signatures were collected using either black or blue ink (No pen brands were taken into consideration), on a white A4 sheet of paper, with eight signature per page. A scanner subsequently digitized the eight signatures, contained on each page, with a resolution in 256 grey levels. Afterwards the images were cut and pasted in rectangular areas of size 256x256 pixels. Sample signature database for 16 persons are shown in Fig.2. A group of 16 persons are selected for 12 specimen signatures which make the total of 16x12=192 signature database.

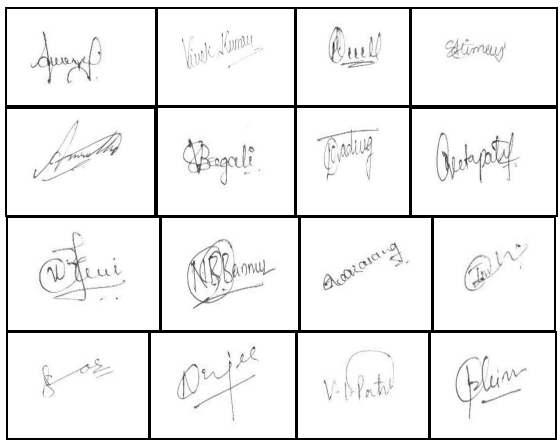

Fig.2. Sample Signature Images Database

Query image

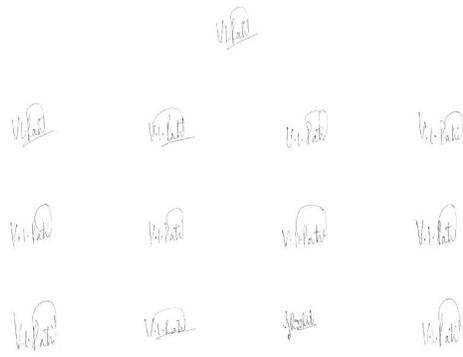

Fig.2. Sample Handwritten Signature Images Database

### B. Retrieval Performance

For each experiment, one image was selected at random as the query image from each writer and thus retrieved images were obtained. For performance evaluation of the signature image retrieval system, it is significant to define a suitable metric. Two metrics are employed in our experiments as follows.

$$Recall = \frac{Number\ of\ relevant\ signatures\ retrieved}{Number\ of\ relevant\ signatures} \quad (5)$$

$$Precision = \frac{Number\ of\ relevant\ signatures\ retrieved}{Number\ of\ signatures\ retrieved} \quad (6)$$

Results correspond to precision and recall rate for a Top1, Top 2, Top 5, Top 8, Top 10, and Top 12. The comparative retrieval performance of the proposed system is shown in Table 1.

**Table1: Average Retrieval Performance**

|  | Discrete wavelet Transform | | Curvelet Transform | |
|---|---|---|---|---|
| **Number of Top matches** | Precision % | Recall % | Precision % | Recall % |
| **Top 1** | 100 | 8 | 100 | 8 |
| **Top 2** | 80 | 12.6 | 96.6 | 15.4 |
| **Top 5** | 66.7 | 28.9 | 92 | 36.7 |
| **Top 8** | 55.8 | 37.2 | 73.3 | 48.5 |
| **Top 10** | 51.3 | 43.4 | 70.7 | 59.0 |
| **Top 12** | 47.8 | 47.5 | 66.04 | 65.2 |

Retrieval performance of the proposed method is compared using DWT transform technique. We evaluated the performance in terms of average rate of retrieving images as function of the number of top retrieved images. Fig.3 shows graph illustrating this comparison between DWT and curvelet transform according to the number of top matches considered for database. From Fig. 3, it is clear that the new method is superior to DWT. To retrieve images from the database those have a similar writing style to the original request. In Fig. 4, retrieval example results are presented in a list of images having a query image.

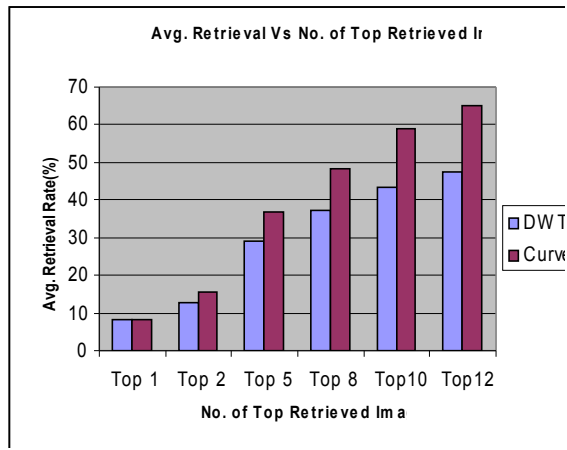

Fig.3. Comparative average retrieval rate using DWT and Curvelet transform

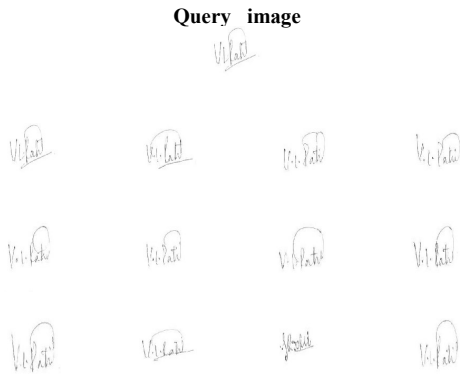

Fig. 4. Sample handwritten signature retrieval example

## V. CONCLUSION

Experimental were conducted for quick for retrieval of offline signature and result are presented. The retrieval performance of the proposed method based on edge correspondence is compared with the retrieval method based on DWT. The proposed method is simple, efficient and outperforms the retrieval system based on curvelet features respect to all parameters (Precision, Recall and Correct retrieval). The proposed approach used curvelet features for extracting details and Canberra distance for comparing features.

AUTHORS PROFILE

M. S. Shirdhonkar completed his B. E., and M.E. from the Department of Computer Science and Engineering, Shivaji University , Kolhapur, India in the years 1994, 2005 respectively. From 1997-2000, he was worked as Lecturer in Computer Science Department at JCE, Institute of Technology , Junner, Maharastra, India. In 2000, he joined as a lecturer in the Department of Computer Science at B. L. D. E' s. Institute of Engineering and Technology, Bijapur, Karnataka, India, where he is presently holding position of Assistant Professor and doing PhD at S.R.T.M. University, Nanded, Maharastra, India. His research interests include image processing, pattern recognition, and document image retrieval. He is a life member of Indian Society for Technical Education and Institute of Engineers.

Manesh Kokare (S'04) was born in Pune, India, in Aug 1972. He received the Diploma in Industrial Electronics Engineering from Board of Technical Examination, Maharashtra, India, in 1990, and B.E. and M. E. Degree in Electronics from Shri Guru Gobind Singhji Institute of Engineering and Technology Nanded, Maharashtra, India, in 1993 and 1999 respectively, and Ph.D. from the Department of Electronics and Electrical Communication Engineering, Indian Institute of Technology, Kharagpur, India, in 2005. Since June1993 to Oct1995, he worked with Industry. From Oct 1995, he started his carrier in academics as a lecturer in the Department of Electronics and Telecommunication Engineering at S. G. G. S. Institute of Engineering and Technology, Nanded, where he is presently holding position of senior lecturer. His research interests include wavelets, image processing, pattern recognition, and Content Based Image Retrieval.…